\documentclass[11pt]{article}
\usepackage{graphicx}
\usepackage{url}
\usepackage{authblk}
\usepackage[a4paper,margin=1in]{geometry}
\usepackage{setspace}
\usepackage{caption}
\usepackage{booktabs}
\usepackage{tabularx}
\usepackage{multirow}
\usepackage{array}
\usepackage{placeins}
\usepackage{xcolor}
\usepackage[hidelinks]{hyperref}
\title{TikStance: A Multimodal and Hierarchical Dataset for Multi-target Stance Analysis in TikTok Political Conversations}
\author[1]{Yazhi Zhang}
\author[2]{Fuqiang Niu}
\author[1]{Bowen Zhang}
\affil[1]{School of Artificial Intelligence, Shenzhen Technology University, Shenzhen, China}
\affil[2]{School of Cyber Science and Technology, University of Science and Technology of China, Hefei, China}
\date{}

\begin{document}
\maketitle

\section*{Abstract}

Political discourse has increasingly moved to short-video platforms, yet computational analysis of such content remains constrained by the scarcity of datasets that jointly preserve audiovisual information and hierarchical conversations. Here we present TikStance, a multimodal and context-aware dataset comprising 161 videos and 13,876 comments from TikTok, designed for stance detection in political discussions. The dataset covers three major political figures in the 2024 U.S. election cycle—Donald Trump, Joe Biden, and Kamala Harris—with content collected between September 2023 and January 2025. Each discussion unit links a host video and its metadata to a parent-linked comment tree, enabling stance analysis within both audiovisual and conversational context. Each item was independently labeled by three annotators using a three-class scheme (Favor, Against, None) for video-to-target and comment-to-target stance; items with disagreement were re-annotated, and the final Krippendorff’s \(\alpha\) reached 0.743, 0.723, and 0.722 for the Trump, Biden, and Harris subsets, respectively. Descriptive analysis further reveals target-dependent differences in stance distributions and conversational depth, with nested replies accounting for 23.3\% of all comments. By combining multi-target coverage, hierarchical conversations, and reliable multi-level human annotations, TikStance supports research in multimodal stance detection, political communication, computational social science, and context-aware natural language processing.

\noindent\textbf{Keywords:} stance detection; TikTok; multimodal dataset; conversational stance; political communication; hierarchical comments

\section{Introduction}
\label{sec:introduction}

Stance detection asks whether a piece of content favors, opposes, or expresses no classifiable position toward a designated target. The target matters: approval of a video, criticism of its speaker, and support for a political candidate are not interchangeable judgments. This relational formulation distinguishes stance from general sentiment and has shaped benchmark tasks since SemEval-2016 and subsequent political stance datasets~\cite{mohammad2016semeval,li2021pstance}. Most early benchmarks package each observation as a self-contained text. That unit is useful, but it leaves out the shared context of a video-hosted discussion.

Short-video conversations make the missing context tangible. A terse reply such as ``exactly,'' ``he did it again,'' or a sequence of emojis may point to speech in the video, an on-screen caption, an editing choice, or an earlier comment. Those cues can agree, conflict, or change as a branch develops. TikTok is also a relevant site for studying this problem: research on the 2024 United States election documents systematic political content patterns on the platform rather than treating it as a neutral window onto public opinion~\cite{ibrahim2026partisan}. The distinction is important. A query-based TikTok collection can reveal properties of the retained sample, but it cannot stand in for the electorate, the platform as a whole, or the causal influence of political events.

Several strands of stance research have moved beyond isolated text. Multi-turn datasets retain reply histories and expose models to contextual dependence. Multimodal benchmarks pair text with images, sampled frames, or transcripts. More recent resources connect political posts to comments, or label comments with respect to a video and an external target. Taken together, these advances show that conversation, multimodality, and video-linked comments are established ideas. They also preserve different slices of the observation unit. Deep conversational datasets remain predominantly textual; video-oriented resources often omit user reply structure; and video-comment collections need not preserve native audiovisual content, complete parent links, or item-level human labels.

The resulting gap is conjunctive rather than absolute. The resources compared here do not combine host-video context, a parent-linked multilevel comment tree, and human judgments for video-to-target and comment-to-target stance in one discussion unit. Building such a unit is not a matter of appending a file to a comment table. Media records must be aligned with platform metadata, missing or deleted parents must be handled without silently flattening branches, and the video- and comment-level stance questions require separate decision rules. Political video and verbatim comments also raise access, licensing, and residual-identifiability concerns. These are construction and governance problems as much as modeling problems.

TikStance addresses this intersection through a video-anchored data model and a co-equal descriptive analysis. One host video, its metadata, retained comments and parent references, and human labels under a common Favor/Against/None vocabulary form a discussion unit. The three target subsets contain 102 videos and 7,676 comments for Trump, 29 videos and 3,177 comments for Biden, and 30 videos and 3,023 comments for Harris, for a total of 161 videos and 13,876 comments. These are post-filter counts for the analyzed sample.

The paper pursues two objectives. First, it specifies an auditable route from political short-video retrieval to video-linked comment trees and video- and comment-level stance labels. Second, it asks what becomes visible when all three targets are described under the same scale, label, and reply-depth definitions. Three contributions follow from those objectives:

\begin{enumerate}
    \item \textbf{A video-grounded hierarchical observation unit.} The TikStance schema is designed to retain host-video context and parent-linked reply structure rather than reduce a discussion to static media proxies or independent comments.
    \item \textbf{Two aligned human stance views.} The annotation design distinguishes the video's stance toward the political target and a comment's stance toward that same target, judged within the video-anchored conversation. These are separate judgments rather than a single label transferred across levels.
    \item \textbf{A three-target descriptive analysis.} Common definitions expose differences in subset size, post-filter stance composition, and conversational depth. The analysis turns those differences into recommendations for per-target reporting, video-level partitioning, contextual ablations, and cross-target evaluation.
\end{enumerate}

The remaining sections follow the data lifecycle. Related Work establishes the scoped feature gap. Dataset Construction, Records, and Validation describes construction, the data model, the current availability status, and quality checks; Data Analysis examines target, annotation-level, and structural patterns. Discussion and Usage develops research questions, reuse guidance, and the boundaries of what the descriptive sample can support. A concluding synthesis and an ethics statement close the paper.

\section{Related Work}
\label{sec:related}

\subsection{Target-specific and political stance benchmarks}

The modern stance-detection formulation treats the target as part of the prediction problem. SemEval-2016 Task 6 established a three-way tweet classification setting over predefined targets~\cite{mohammad2016semeval}. P-Stance later concentrated on political targets, including Donald Trump and Joe Biden, and supported cross-target comparison within a common label space~\cite{li2021pstance}. Such resources remain valuable because they isolate the semantic relation between an utterance and a target. Their usual observation unit, however, is an individual post. It does not include a host video or a branch of replies as evidence available to the annotator or model.

Target diversity has become a research question in its own right. ZS-CSD constructs conversational samples from Weibo discussion trees across a much larger target inventory and explicitly studies zero-shot transfer~\cite{ding2025zscsd}. That work makes two points relevant here: target-specific lexical regularities can be brittle, and an aggregate score can conceal substantial target variation. TikStance does not claim comparable target breadth. Its narrower three-target design instead permits a controlled descriptive comparison within one political setting, provided each target is reported separately and the unequal subset sizes remain visible.

\subsection{Conversational stance detection}

Conversation-oriented datasets restore context that single-message benchmarks discard. MT-CSD contains human-annotated English Reddit instances with multi-turn histories, while C-MTCSD extends deep conversational stance detection to Chinese Weibo discussions~\cite{niu2024challenge,niu2025cmtcsd}. ZS-CSD likewise derives examples from discussion trees, including deep contexts~\cite{ding2025zscsd}. These resources establish that reply history and multilevel conversation are not new contributions by themselves. They also show why a comment should not automatically be treated as independent of its ancestors.

Recent work has begun to bridge conversation and modality. Cause-CSD augments multi-turn discussions with embedded images and stance-cause links~\cite{niu2026causecsd}. Its existence makes the blanket claim that conversational stance datasets are text-only untenable. The relevant distinction is narrower: image-grounded dialogue is not the same data object as a reply tree anchored to a native short video with temporal visual and acoustic content. TikStance therefore does not position hierarchy or multimodality as isolated novelties. It asks whether those properties can be joined at the level of a reusable discussion unit.

\subsection{Multimodal and video-grounded stance resources}

Multimodal stance benchmarks have expanded the evidence associated with a target-bearing post. MMSD pairs text and images across several domains and targets, using human annotation; when a source post contains a video or GIF, the resource represents it through a first-frame proxy~\cite{liang2024multimodal}. MultiClimate samples frames from climate-change videos and aligns them with transcripts for stance labeling~\cite{wang2024multiclimate}. The latter is a direct video-content comparator, but its units are frame--transcript pairs rather than user comment trees, and it does not model native audio as a separate input. Both datasets clarify the value of visual evidence without claiming to preserve a hierarchical audience conversation.

StanceGen2024 is especially close in political period and post--comment structure. It pairs multimodal posts with comments from the 2024 United States election, but converts video and GIF attachments to their first frame and does not report a parent-linked reply hierarchy~\cite{wang2025stancegen}. Its hybrid label pipeline also differs from a design based on independent human judgments at every annotation level. The comparison matters because it removes an overly broad novelty claim: election-focused multimodal post--comment stance data already exist.

DIVERSE is the nearest video-comment resource. It labels YouTube comments with respect to both the associated video and the U.S. Army through a data-programming pipeline~\cite{cruickshank2024diverse}. Replies are collected and a root conversation identifier is documented, but the published field description does not provide the parent pointers needed to reconstruct a complete multilevel tree. Nor is a native full-video object documented as the modeling input. DIVERSE thus precedes TikStance in video-linked comment stance; TikStance's proposed distinction lies in joining native-video context, explicit parent relations, a video-level target label, and independent human comment-level judgments.

\subsection{Scoped position of TikStance}

The relevant resources form three overlapping groups: deep textual conversations, multimodal or video-derived stance without user discussion structure, and video- or post-linked comments without the complete TikStance design. Cause-CSD closes part of the modality--conversation divide with images. StanceGen2024 and DIVERSE close part of the post--response divide. MT-CSD, C-MTCSD, and ZS-CSD provide stronger depth or target coverage than TikStance claims. No component is new on its own.

TikStance is positioned at the intersection of these groups through a schema that joins host-video context, parent-linked multilevel comments, and human stance labels at both the video and the comment level toward a common designated target. This is a feature-combination claim about the resources discussed above, not a claim that conversation, multimodality, or multi-level stance annotation is individually novel.

\section{Dataset Construction, Records, and Validation}
\label{sec:dataset}
\label{sec:methods}

\begin{figure}[htbp]
    \centering
    \includegraphics[width=0.54\linewidth]{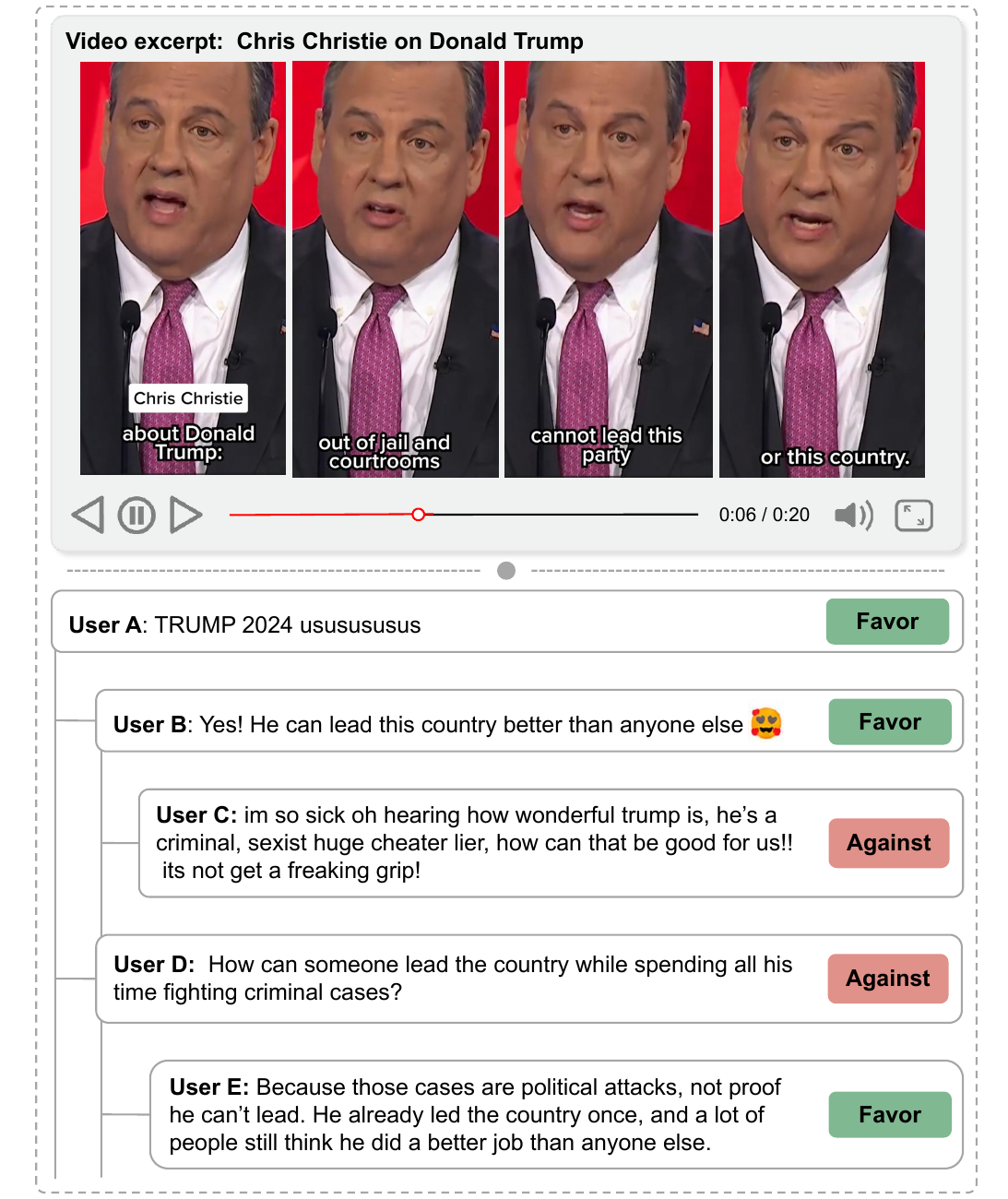}
    \caption{\textbf{Illustrative TikStance discussion unit.} Sampled host-video frames are shown with comments expressing different stances toward Donald Trump. Account identifiers are replaced by generic user labels. The example illustrates why video and conversational context may be required; it is not evidence for a population-level or causal claim.}
    \label{fig:conversation-example}
\end{figure}

The discussion unit is illustrated in Figure~\ref{fig:conversation-example}. It places the host-video context and a set of target-directed comment judgments in one view; the quantitative analyses below rely on aggregate source tables rather than on this single example.

\subsection{Study Scope and Data Lifecycle}
\label{sec:study_scope}

TikStance defines a video-anchored \emph{discussion unit}: one TikTok host video assigned to a designated political target, the associated video metadata, and the retained comments connected through parent--reply relations. We focused on English-language discussions of Donald Trump, Joe Biden, and Kamala Harris collected between September 2023 and 20 January 2025. The observation window spans the late-2023 pre-primary period, the 2024 primaries, the Trump--Biden and Trump--Harris campaign phases, Election Day, and the period through the presidential inauguration. These phases provide political context; they are not used as causal treatments or as evidence of exhaustive platform coverage.

Construction proceeded from hashtag-based discovery to multimodal screening, comment retrieval, reply-tree reconstruction, content cleaning, independent human annotation, label aggregation, and record assembly (Table~\ref{tab:construction_lifecycle}). The engagement thresholds and targeted queries intentionally favor visible political discussions. TikStance is therefore a non-probability research sample rather than a census of TikTok or a sample of voters.

\begin{table}[htbp]
    \centering
    \small
    \caption{TikStance construction lifecycle. Counts are shown where they are available in the study records; the intermediate stages were applied sequentially.}
    \label{tab:construction_lifecycle}
    \begin{tabularx}{\textwidth}{p{2.8cm}X p{3.0cm}}
        \toprule
        \textbf{Stage} & \textbf{Procedure} & \textbf{Recorded output} \\
        \midrule
        Candidate discovery & Election- and candidate-related hashtag queries followed by a discovery-stage threshold of more than 5,000 displayed likes and comments. & 2,397 candidate videos \\
        Video screening & Multimodal relevance review, designation of one political target, and exclusion of videos with fewer than 100 retrieved comments. & Retained discussion units \\
        Comment processing & Comment and reply retrieval, parent-link reconstruction, English-language filtering, spam and duplicate removal, and contextual cleaning. & Retained comment trees \\
        Human annotation & Independent triad judgments for video relevance and video- and comment-level stance, followed by re-annotation of inconsistent items and majority aggregation. & Video- and comment-level labels \\
        Dataset assembly & Organization by target and video identifier, with direct account identifiers removed from the research records. & 161 videos and 13,876 comments \\
        \bottomrule
    \end{tabularx}
\end{table}

\subsection{Query Design and Candidate Retrieval}
\label{sec:retrieval}

Public TikTok videos were located with a manually curated lexicon of election- and candidate-related hashtags. The lexicon covered general election discourse, candidate names and campaign slogans, party or ideological terms, salient campaign events, and voting or result terms (Table~\ref{tab:hashtags}). Directional hashtags such as \#nevertrump and \#voteblue were retained because the retrieval goal was coverage of contested political discussion rather than a sentiment-neutral keyword sample.

\begin{table}[htbp]
    \centering
    \small
    \caption{Hashtag categories and representative retrieval terms used in candidate discovery.}
    \label{tab:hashtags}
    \begin{tabularx}{\textwidth}{p{3.0cm}X}
        \toprule
        \textbf{Category} & \textbf{Representative hashtags} \\
        \midrule
        General election & \#election2024, \#uselection, \#presidentialelection, \#vote2024, \#electionday2024, \#politics \\
        Trump-related & \#trump, \#donaldtrump, \#trump2024, \#votetrump, \#maga, \#makeamericagreatagain, \#nevertrump \\
        Biden-related & \#biden, \#joebiden, \#biden2024, \#trumpvsbiden, \#bidentrumpdebate, \#bidendebate \\
        Harris-related & \#kamalaharris, \#harris2024, \#trumpvsharris, \#harrisvstrump, \#voteharris \\
        Party or camp & \#republican, \#democrat, \#gop, \#liberal, \#conservative, \#voteblue, \#votered \\
        Key events & \#presidentialdebate, \#debate2024, \#rnc2024, \#dnc2024, \#trumprally, \#trumpassassinationattempt \\
        Voting or results & \#earlyvoting, \#electionday, \#electionresults, \#trumpwins, \#presidentelecttrump \\
        \bottomrule
    \end{tabularx}
\end{table}

At discovery, videos were required to display more than 5,000 likes and more than 5,000 comments, yielding 2,397 candidates. A second threshold was applied after retrieval: videos with fewer than 100 stored comments were removed. The two thresholds refer to different stages of the pipeline---platform-facing engagement during discovery and comments recovered for analysis---and were not treated as interchangeable measurements. For each candidate, the collection process requested the host video, platform identifier, title or description and hashtags, share URL, comment text and metadata, and reply-related identifiers.

\subsection{Video Screening and Designated-Target Assignment}
\label{sec:video_screening}

Reviewers screened candidate videos for substantive relevance to Trump, Biden, or Harris. They considered spoken content, visual scenes, on-screen captions, banners, political symbols, and editing choices rather than relying on titles alone. Videos centered on unrelated figures or generic topics that did not support a target-specific judgment were excluded. Each retained discussion unit was assigned one designated target, fixing the object of all subsequent video- and comment-level target-stance judgments. A relevant video could receive the label \texttt{None} when it concerned the target without expressing a discernible Favor or Against orientation.

\subsection{Comment Retrieval and Reply-Tree Reconstruction}
\label{sec:tree_reconstruction}

Three identifiers define the conversation hierarchy. The field \texttt{cid} identifies a comment. For a top-level comment, \texttt{reply\_id} is zero; for a reply it identifies the top-level root of the thread. The field \texttt{reply\_to\_reply\_id} is zero for a direct response to that root and otherwise stores the immediate-parent reply. Depth is the number of parent edges between a comment and its top-level root: top-level comments have depth 0, direct replies depth 1, and deeper replies successively larger values.

This representation preserves both the root thread and the immediate parent needed to reconstruct a branch. It also prevents a frequent modeling error: treating every reply as an independent response to the host video. Analyses in this paper use the reported depth marginals. Record-level reuse should additionally enforce identifier uniqueness, same-video parentage, parent existence, acyclicity, and agreement between stored and recomputed depth.

\subsection{Cleaning, Language Filtering, and Deduplication}
\label{sec:cleaning}

Cleaning was performed from the thread level downward so that a retained reply did not lose context solely because one of its ancestors was filtered. Threads unrelated to the designated target were removed. Within retained threads, the process excluded non-English, emoji-only, duplicate, spam, and contextually uninterpretable comments, while preserving ancestors needed to interpret a retained descendant. Records were deleted rather than linguistically rewritten.

Target relevance and the stance label \texttt{None} are conceptually separate. \texttt{None} denotes a relevant item for which the annotation codebook assigns neither Favor nor Against; it is not a container for missing context, unusable text, or an unrelated topic. The empirical class proportions reported below are consequently post-filter compositions of the retained sample. They may reflect both political content and the retrieval and cleaning rules.

\subsection{Annotation Concepts and Codebook}
\label{sec:annotation_scheme}

The annotation design separates relevance from stance and distinguishes two stance relations. At the video level, annotators judge whether the host video is related to the designated target and, when related, whether it favors, opposes, or takes no classifiable position toward that target. At the comment level, annotators judge the comment's stance toward the designated target within its conversational context.

All stance tasks use \texttt{Favor}, \texttt{Against}, and \texttt{None}. \texttt{Favor} indicates support for or positive framing of the relevant object, \texttt{Against} indicates opposition or negative framing, and \texttt{None} indicates relevant content without a discernible position in either direction. The stance object remains explicit: both judgments concern the designated political target, so \texttt{stance\_target} is neither a copy of the video-level label nor a label for stance toward the immediate parent comment.

\begin{table}[htbp]
    \centering
    \small
    \caption{Human annotation tasks and their stance objects.}
    \label{tab:annotation_tasks}
    \begin{tabularx}{\textwidth}{p{2.0cm}p{3.3cm}p{3.0cm}X}
        \toprule
        \textbf{Level} & \textbf{Judgment} & \textbf{Allowed values} & \textbf{Object of the judgment} \\
        \midrule
        Video & Relevance & \texttt{related}, \texttt{unrelated} & Whether the host video is substantively related to the designated target. \\
        Video & Video-level stance & \texttt{Favor}, \texttt{Against}, \texttt{None} & The host video's stance toward the designated target. \\
        Comment & \texttt{stance\_target} & \texttt{Favor}, \texttt{Against}, \texttt{None} & The comment's stance toward the designated target. \\
        \bottomrule
    \end{tabularx}
\end{table}

Annotators could consult the host video, the parent comment, and the preceding reply chain to resolve pronouns, ellipsis, sarcasm, quotation, and other context-dependent expressions. This common context policy is central to the dataset design: labels are judgments about a target-conditioned discussion unit, not only about an isolated text string.

\subsection{Annotators, Quality Control, and Aggregation}
\label{sec:annotators}
\label{sec:label_aggregation}

Twelve annotators participated. Each had at least a bachelor's degree, working English proficiency, and familiarity with social-media content. A shared training session covered task definitions, stance objects, decision rules, and boundary cases including sarcasm, quotations, emojis, and partisan slogans. Annotators were organized into fixed groups of three; the members of each group labeled the same assigned items independently and without discussion during initial annotation.

Binary video-relevance labels were combined by majority vote. For the three-way stance tasks, the three annotators in a group labeled each item independently; items on which the annotators disagreed were returned for re-annotation, and residual disagreements were resolved by the label selected by at least two annotators. Inter-annotator reliability was then quantified with Krippendorff's \(\alpha\), which is suitable for coding reliability when the measurement scale, units, raters, and missing-data treatment are specified~\cite{artstein2008inter,hayes2007answering}. The final per-target coefficients are reported in Section~\ref{sec:annotation_validation}.

\subsection{Dataset Assembly and De-identification}
\label{sec:assembly}

The documented schema organizes discussion units first by designated target and then by \texttt{video\_id}. That key is intended to link the host-video record, video metadata, video-level annotation, and comments. Comment identifiers and parent references encode the tree, while separate fields retain video-to-target and comment-to-target stance. Usernames, platform user identifiers, avatar links, and profile metadata were removed during preparation. Verbatim text, timestamps, and audiovisual content may nevertheless remain indirectly identifying, a risk addressed in the Ethics Statement.

\subsection{Data Records}
\label{sec:data_records}

\subsubsection{Record Organization}
\label{sec:record_access}

The intended record layout places each discussion unit under a target and video identifier. A host-video object and \texttt{video\_meta.json} describe the video; \texttt{comments.json} contains the retained comment nodes, parent relations, depth, and comment-level labels. This layout is designed to keep media, metadata, conversation structure, and annotations aligned without duplicating the host video for every comment. If media redistribution is restricted, the same identifier-level organization can support controlled media access while leaving metadata and derived records under separately stated terms.

TikStance is not yet associated with a public repository, persistent identifier, or finalized data license. The present article therefore documents the construction, schema, and aggregate characterization without claiming an openly downloadable release. A public version will need to state precisely which media, text, metadata, and identifiers can be distributed, and to distinguish data terms from the software license~\cite{wilkinson2016fair,gebru2021datasheets}.

\subsubsection{Field Dictionary and Join Semantics}
\label{sec:field_dictionary}

Table~\ref{tab:data_fields} defines the fields used by the video and comment records. The cross-file key is \texttt{video\_id}; within the comment file, \texttt{cid} identifies a node and the two reply fields encode its root and immediate parent. Zero is a structural sentinel in the reply fields and must not be conflated with a missing stance label.

\begin{table}[htbp]
    \centering
    \fontsize{7.3}{8.2}\selectfont
    \caption{TikStance record fields and their semantics.}
    \label{tab:data_fields}
    \begin{tabularx}{\textwidth}{p{3.0cm}p{1.7cm}X p{3.2cm}}
        \toprule
        \textbf{File and field} & \textbf{Type} & \textbf{Meaning or allowed values} & \textbf{Origin} \\
        \midrule
        \multicolumn{4}{l}{\textit{video\_meta.json}} \\
        \midrule
        \texttt{video\_id} & String & Discussion-unit key and foreign key for comments. & Platform identifier \\
        \texttt{title} & String & Video title or description, including hashtags where present. & Platform metadata \\
        \texttt{duration} & Float & Video duration in seconds. & Media metadata \\
        \texttt{publish\_time} & Integer & UTC timestamp of video publication. & Platform metadata \\
        \texttt{target} & Categorical & \texttt{Trump}, \texttt{Biden}, or \texttt{Harris}. & Screening assignment \\
        \texttt{relevance} & Categorical & \texttt{related} or \texttt{unrelated}. & Human annotation \\
        \texttt{video\_stance} & Categorical & \texttt{Favor}, \texttt{Against}, or \texttt{None} toward the target. & Human annotation \\
        \midrule
        \multicolumn{4}{l}{\textit{comments.json}} \\
        \midrule
        \texttt{cid} & String & Comment identifier and node key. & Platform identifier \\
        \texttt{video\_id} & String & Foreign key to the host-video discussion unit. & Platform identifier \\
        \texttt{text} & String & Retained comment text. & Platform content \\
        \texttt{create\_time} & Integer & UTC timestamp of comment publication. & Platform metadata \\
        \texttt{digg\_count} & Integer & Comment like count at collection. & Platform metadata \\
        \texttt{reply\_id} & String & Zero for a top-level comment; otherwise the top-level root identifier. & Platform relation \\
        \texttt{reply\_to\_reply\_id} & String & Zero for a direct root reply; otherwise the immediate-parent identifier. & Platform relation \\
        \texttt{depth} & Integer & Number of parent edges from the comment to its top-level root. & Derived \\
        \texttt{stance\_target} & Categorical & \texttt{Favor}, \texttt{Against}, or \texttt{None} toward the target. & Human annotation \\
        \bottomrule
    \end{tabularx}
\end{table}

The aggregate tables available for the present analysis cover \texttt{video\_stance}, \texttt{stance\_target}, and reply depth. Relations between the video- and comment-level stance views require item-level joins rather than marginal totals, so Section~\ref{sec:data-analysis} does not infer them.

\subsubsection{Versioning and Access Status}
\label{sec:versioning}

No versioned public deposit accompanies this manuscript. Before external release, a frozen manifest, machine-readable schema, checksums, change log, access terms, and a withdrawal procedure should identify the exact research objects covered by a citation. This status does not affect the arithmetic analyses below, which operate on the target-level count tables, but it limits claims about public availability and file-level completeness.

\subsection{Technical Validation}
\label{sec:validation}

\label{sec:validation_strategy}

Validation was considered at three levels: annotation reliability, consistency of the aggregate tables, and integrity of the linked record and media objects. These levels have different denominators and should not be collapsed. Agreement among coders does not validate a parent link; a closed aggregate total does not prove that every media file is decodable; and model accuracy would not substitute for either check.

\subsubsection{Annotation Reliability}
\label{sec:annotation_validation}

After re-annotation of inconsistent items, the final inter-annotator reliability reached Krippendorff's \(\alpha=0.743\) for Trump, \(\alpha=0.723\) for Biden, and \(\alpha=0.722\) for Harris (Table~\ref{tab:reliability_status}). All three per-target coefficients exceed the conventional 0.7 threshold for acceptable coding reliability~\cite{artstein2008inter,hayes2007answering}. The coefficients are reported per target rather than pooled, so that the reliability of each subset remains visible alongside its unequal size.

\begin{table}[htbp]
    \centering
    \small
    \caption{Per-target inter-annotator reliability after re-annotation of inconsistent items.}
    \label{tab:reliability_status}
    \begin{tabularx}{\textwidth}{p{2.5cm}p{2.7cm}X}
        \toprule
        \textbf{Target} & \textbf{Reported \(\alpha\)} & \textbf{Interpretation in this paper} \\
        \midrule
        Trump & 0.743 & Per-target reliability of the aggregated stance labels. \\
        Biden & 0.723 & Per-target reliability of the aggregated stance labels. \\
        Harris & 0.722 & Per-target reliability of the aggregated stance labels. \\
        \bottomrule
    \end{tabularx}
\end{table}

\subsubsection{Aggregate Consistency and Coverage}
\label{sec:coverage_validation}

Arithmetic closure was checked across all aggregate count tables. Target totals sum to 161 videos and 13,876 comments. Within every target, the video stance counts equal the video total, the comment-to-target stance counts equal the comment total, and the five depth groups equal the same comment total. Across targets, the stance totals are 70/52/39 videos and 5,035/5,149/3,692 comments for Favor/Against/None. The depth totals are 10,647/2,113/457/276/383 for depths 0/1/2/3/4+. These checks support the descriptive denominators used in Section~\ref{sec:data-analysis}.

Coverage is heterogeneous. Trump contributes most videos and comments; each target nevertheless contains all three target-stance labels at both annotation levels and observations in every reported depth group. The temporal graphics were excluded because their denominators, timestamp proxies, and label-inclusion rules conflict: the Trump video bars total 99 rather than 102, while a Harris panel uses 31 rather than 30 videos and 3,013 rather than 3,023 comments. Excluding those panels prevents incompatible time definitions from being merged into a single series.

\subsubsection{Record and Media Integrity}
\label{sec:structural_validation}
\label{sec:media_validation}

The record model supports deterministic checks for JSON parsing, key uniqueness, comment-to-video joins, parent existence, same-video parentage, cycles, and stored-versus-recomputed depth. Media validation similarly requires a manifest, checksum comparison, container parsing, and decodability checks. Item-level outputs from these tests are not available for the present study, so empirical validation is restricted to annotation summaries and aggregate consistency. We consequently interpret depth counts as structural marginals rather than as proof that every reconstructed tree is complete.

\section{Data Analysis}
\label{sec:data-analysis}

The analyzed TikStance sample contains 161 videos and 13,876 retained comments distributed over Trump, Biden, and Harris discussion units (Table~\ref{tab:analysis-inventory}). Trump contributes 102 videos and 7,676 comments, whereas Biden and Harris contribute 29/3,177 and 30/3,023, respectively. These counts characterize the retained sample and do not estimate candidate popularity, platform exposure, or public opinion. The unequal target sizes make target-stratified reporting and video-grouped data partitions more informative than a single pooled score.

\subsection{Cross-target inventory and observation density}
\label{subsec:analysis-inventory}

The Trump subset accounts for 63.4\% of the videos and 55.3\% of the comments, compared with 18.0\%/22.9\% for Biden and 18.6\%/21.8\% for Harris. Retained comments per video therefore differ in the opposite direction to raw subset size: 75.3 for Trump, 109.6 for Biden, and 100.8 for Harris, with an overall mean of 86.2. Because these ratios are calculated after retrieval and filtering, they are observation densities rather than engagement rates. Analyses that reuse TikStance should consequently report both video and comment denominators and should not allow the larger Trump subset to define the apparent performance of a multi-target model.

\begin{table}[htbp]
\centering
\caption{\textbf{Cross-target inventory of the analyzed sample.} Video and comment shares use 161 videos and 13,876 comments as their respective denominators. Comments per video and nested-reply values describe retained records after filtering; they are not platform engagement estimates.}
\label{tab:analysis-inventory}
\small
\resizebox{\linewidth}{!}{%
\begin{tabular}{lrrrrrr}
\toprule
Target & Videos, $n$ & Video share & Comments, $n$ & Comment share & Comments/video & Nested, $n$ (\%) \\
\midrule
Trump & 102 & 63.4\% & 7,676 & 55.3\% & 75.3 & 1,999 (26.0\%) \\
Biden & 29 & 18.0\% & 3,177 & 22.9\% & 109.6 & 731 (23.0\%) \\
Harris & 30 & 18.6\% & 3,023 & 21.8\% & 100.8 & 499 (16.5\%) \\
\midrule
Total & 161 & 100.0\% & 13,876 & 100.0\% & 86.2 & 3,229 (23.3\%) \\
\bottomrule
\end{tabular}%
}
\end{table}

\subsection{Stance composition across targets and annotation levels}
\label{subsec:analysis-stance}

At the video level, the largest class differs across the three retained target subsets: Favor accounts for 51 of 102 Trump videos (50.0\%), Against for 16 of 29 Biden videos (55.2\%), and Favor and Against each account for 13 of 30 Harris videos (43.3\%). All three labels occur for every target, but their frequencies are not uniform and the Biden and Harris video denominators are small. These within-sample class priors support per-target and per-class evaluation, with any weighting or resampling estimated from training partitions rather than from the complete dataset.

At the comment level, Trump has a Favor plurality of 3,570 of 7,676 comments (46.5\%), Biden has a None majority of 1,669 of 3,177 (52.5\%), and Harris is close to parity between Favor (1,108; 36.7\%) and Against (1,092; 36.1\%). Pooled over targets, Against is the largest comment class by only 114 records: 5,149 of 13,876 (37.1\%) versus 5,035 (36.3\%) Favor, while 3,692 (26.6\%) are None. The pooled distribution is weighted by the unequal target sizes and therefore should not be interpreted as a platform-wide stance estimate. For reuse, macro-averaged class metrics should be accompanied by target-specific results so that the pooled total does not conceal the Biden None concentration or the Harris near-parity pattern.

\begin{table}[htbp]
\centering
\caption{\textbf{Target-directed stance composition by annotation level.} Each cell reports count and within-target percentage. Comment rows refer to comment-to-target stance. Percentages were recomputed from the displayed integer counts and rounded to one decimal place; no confidence intervals or inferential tests are implied.}
\label{tab:analysis-stance}
\small
\begin{tabular}{llrrrr}
\toprule
Target & Unit & Favor, $n$ (\%) & Against, $n$ (\%) & None, $n$ (\%) & Total \\
\midrule
\multirow{2}{*}{Trump}
 & Video & 51 (50.0\%) & 23 (22.5\%) & 28 (27.5\%) & 102 \\
 & Comment & 3,570 (46.5\%) & 2,906 (37.9\%) & 1,200 (15.6\%) & 7,676 \\
\addlinespace
\multirow{2}{*}{Biden}
 & Video & 6 (20.7\%) & 16 (55.2\%) & 7 (24.1\%) & 29 \\
 & Comment & 357 (11.2\%) & 1,151 (36.2\%) & 1,669 (52.5\%) & 3,177 \\
\addlinespace
\multirow{2}{*}{Harris}
 & Video & 13 (43.3\%) & 13 (43.3\%) & 4 (13.3\%) & 30 \\
 & Comment & 1,108 (36.7\%) & 1,092 (36.1\%) & 823 (27.2\%) & 3,023 \\
\midrule
\multirow{2}{*}{Total}
 & Video & 70 (43.5\%) & 52 (32.3\%) & 39 (24.2\%) & 161 \\
 & Comment & 5,035 (36.3\%) & 5,149 (37.1\%) & 3,692 (26.6\%) & 13,876 \\
\bottomrule
\end{tabular}
\end{table}

\begin{table}[htbp]
\centering
\caption{\textbf{Comment-minus-video marginal stance contrast.} Values are percentage-point differences calculated from unrounded count ratios. They compare marginal compositions of different units and are not paired transitions or effect estimates.}
\label{tab:analysis-cross-level}
\small
\begin{tabular}{lrrr}
\toprule
Target & Comment--video Favor & Comment--video Against & Comment--video None \\
\midrule
Trump & $-3.5$ & $+15.3$ & $-11.8$ \\
Biden & $-9.5$ & $-18.9$ & $+28.4$ \\
Harris & $-6.7$ & $-7.2$ & $+13.9$ \\
\midrule
Total & $-7.2$ & $+4.8$ & $+2.4$ \\
\bottomrule
\end{tabular}
\end{table}

\begin{figure}[h]
\centering
\begin{minipage}[t]{0.48\linewidth}
\centering
\textbf{a}\par\smallskip
\includegraphics[width=\linewidth]{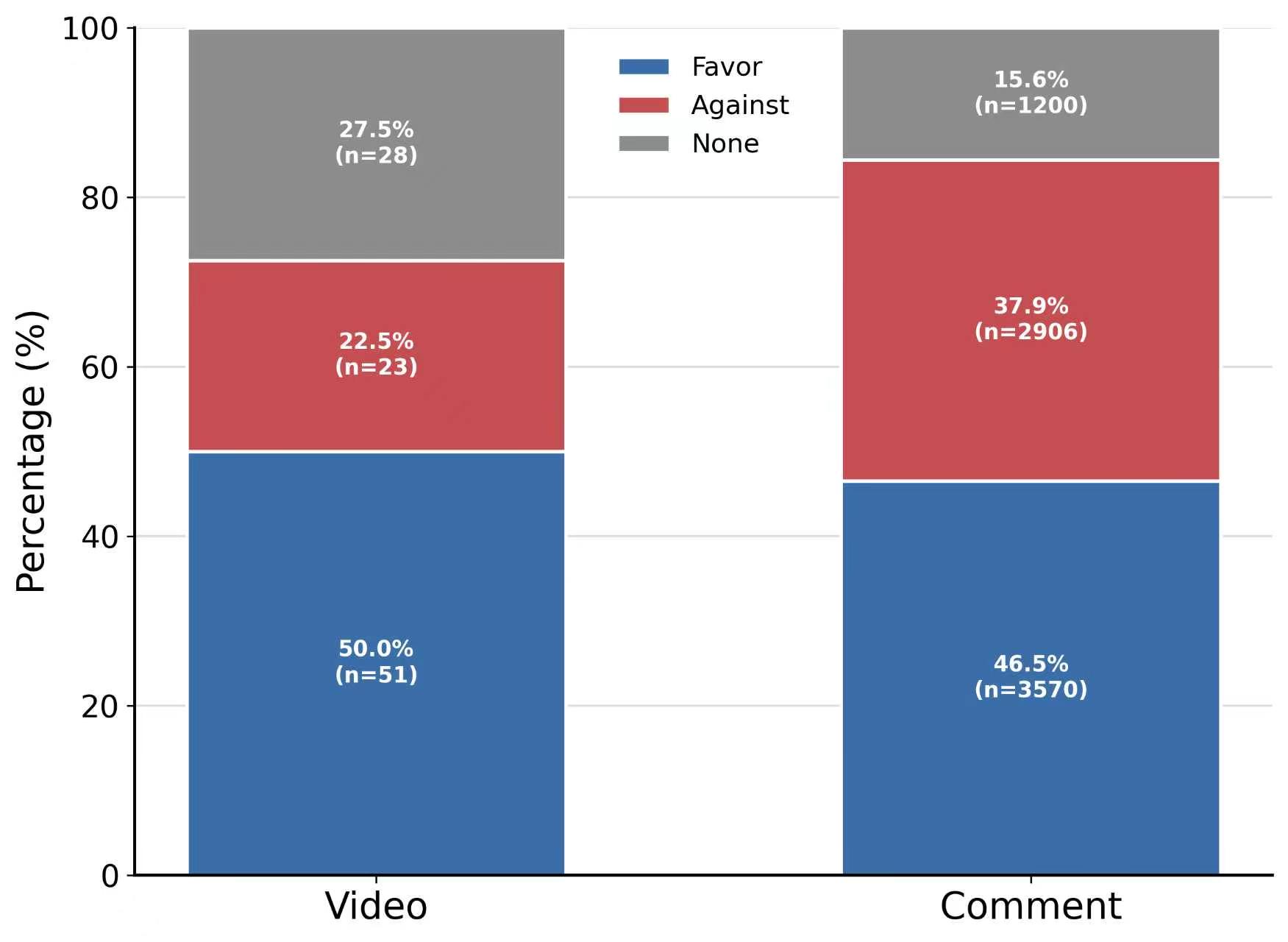}
\end{minipage}\hfill
\begin{minipage}[t]{0.48\linewidth}
\centering
\textbf{b}\par\smallskip
\includegraphics[width=\linewidth]{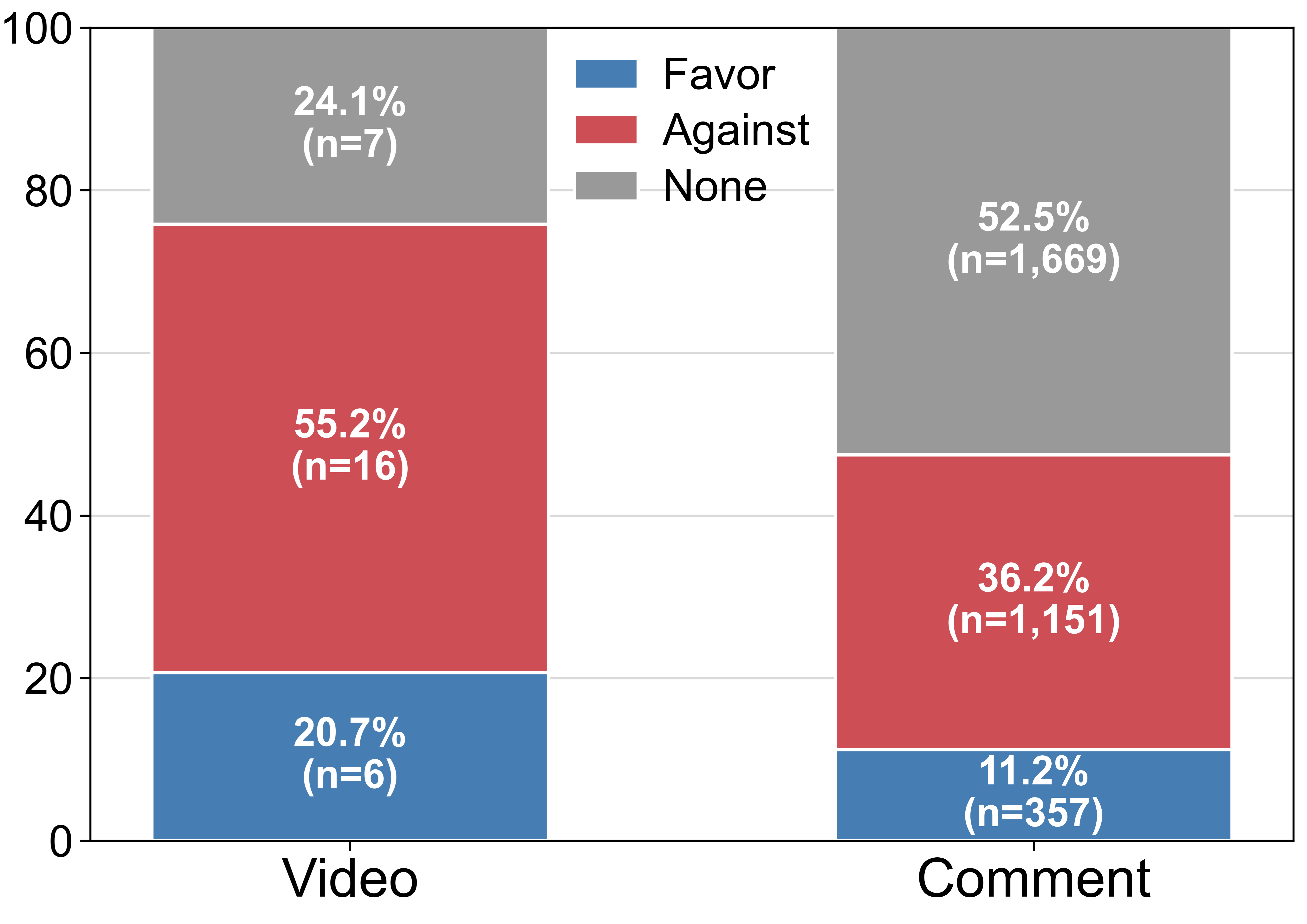}
\end{minipage}

\vspace{0.7em}
\begin{minipage}[t]{0.48\linewidth}
\centering
\textbf{c}\par\smallskip
\includegraphics[width=\linewidth]{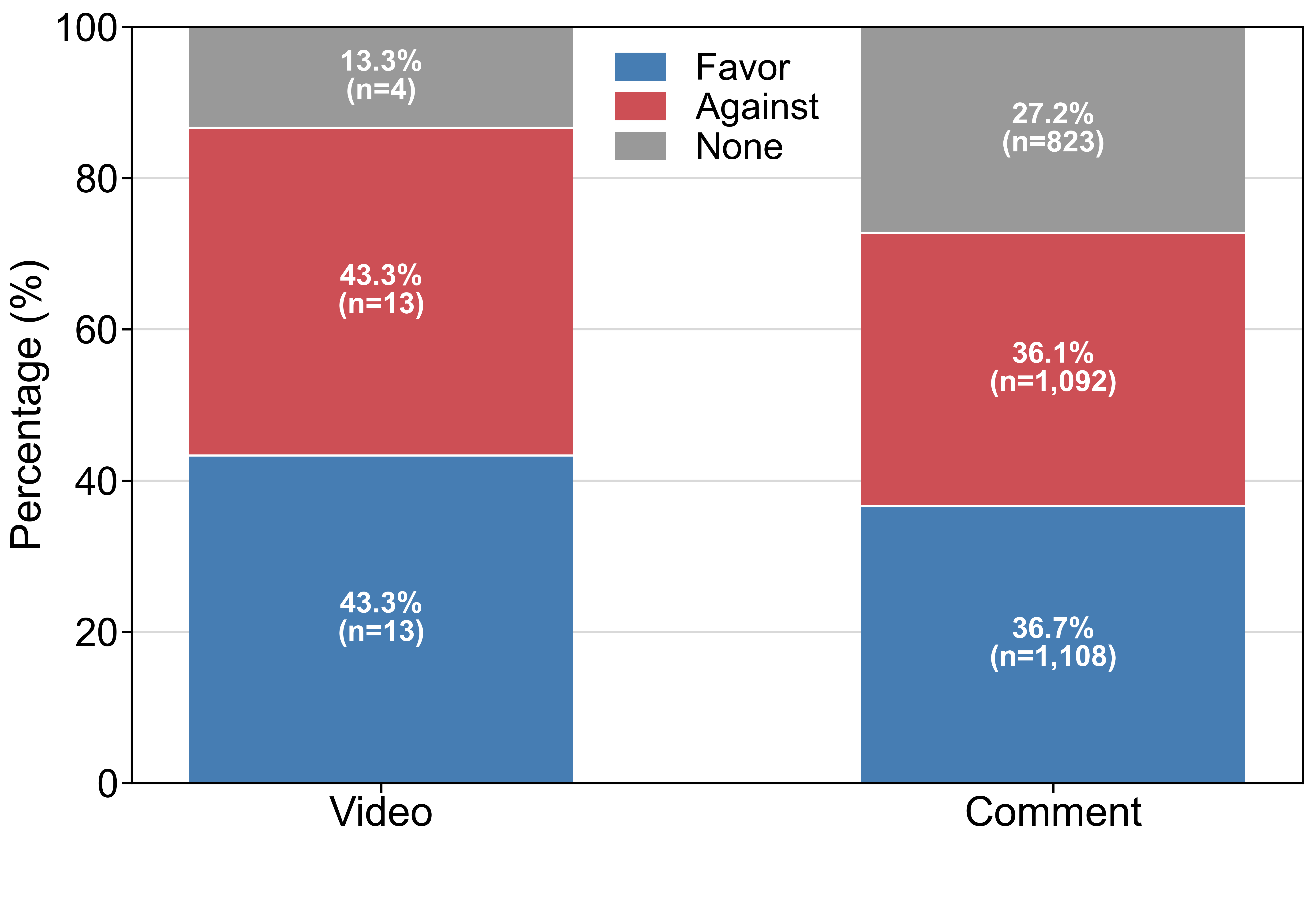}
\end{minipage}
\caption{\textbf{Video- and comment-level stance composition.} Panels show (a) Trump, (b) Biden, and (c) Harris.}
\label{fig:analysis-stance}
\end{figure}

Figure~\ref{fig:analysis-stance} visualizes the target-specific stance counts in Table~\ref{tab:analysis-stance}. It communicates target-conditioned class composition rather than inferential separation among targets.

The comment-minus-video differences are also target dependent (Table~\ref{tab:analysis-cross-level}). Against increases by 15.3 percentage points for Trump but decreases by 18.9 points for Biden and 7.2 points for Harris; None decreases by 11.8 points for Trump but increases by 28.4 and 13.9 points for Biden and Harris. Favor is lower at the comment level for all three targets, by 3.5, 9.5, and 6.7 points, respectively. Video labels and comment labels describe different units with different denominators, and comments are clustered within videos, so these contrasts are compositional differences rather than paired transitions or creator-to-audience effects. Models should therefore treat video-to-target and comment-to-target stance as distinct prediction settings and should not collapse their label distributions into one task prior.

\subsection{Reply depth and retained conversational structure}
\label{subsec:analysis-depth}

Of 13,876 retained comments, 10,647 (76.7\%) are top-level comments at depth 0 and 3,229 (23.3\%) are nested replies. The nested set comprises 2,113 comments at depth 1 (15.2\% of all comments), 457 at depth 2 (3.3\%), 276 at depth 3 (2.0\%), and 383 at depth 4 or deeper (2.8\%). Thus, nearly one quarter of the comments have at least one conversational ancestor, although the grouped 4+ bin does not reveal the maximum tree depth. Conversational evaluations should preserve parent and ancestor context and split by the anchoring video rather than distributing descendants of the same discussion across training and evaluation sets.

Nested-reply prevalence ranges from 1,999 of 7,676 comments (26.0\%) for Trump to 731 of 3,177 (23.0\%) for Biden and 499 of 3,023 (16.5\%) for Harris, a maximum descriptive spread of 9.5 percentage points. Depth-4-or-deeper comments remain sparse for every target---237 (3.1\%), 75 (2.4\%), and 71 (2.3\%), respectively---so target and depth are jointly imbalanced. These differences may arise from retention, collection windows, or target-specific sampling and do not establish that one candidate's audience is intrinsically more conversational. Depth-stratified results should retain raw cell sizes, especially for the deepest bins.

\begin{table}[htbp]
\centering
\caption{\textbf{Reply-depth distribution by target.} Cells report comment count and percentage of the corresponding target total. Depth 0 denotes a top-level comment; 4+ combines every observed depth of at least four and therefore does not specify a maximum depth. Nested combines depths 1--4+. Aggregate depth counts reconcile with each target's comment denominator; no stance-by-depth contingency table is used.}
\label{tab:analysis-depth}
\small
\resizebox{\linewidth}{!}{%
\begin{tabular}{lrrrrrrr}
\toprule
Target & Depth 0, $n$ (\%) & Depth 1, $n$ (\%) & Depth 2, $n$ (\%) & Depth 3, $n$ (\%) & Depth 4+, $n$ (\%) & Nested, $n$ (\%) & Total \\
\midrule
Trump & 5,677 (74.0\%) & 1,310 (17.1\%) & 285 (3.7\%) & 167 (2.2\%) & 237 (3.1\%) & 1,999 (26.0\%) & 7,676 \\
Biden & 2,446 (77.0\%) & 501 (15.8\%) & 96 (3.0\%) & 59 (1.9\%) & 75 (2.4\%) & 731 (23.0\%) & 3,177 \\
Harris & 2,524 (83.5\%) & 302 (10.0\%) & 76 (2.5\%) & 50 (1.7\%) & 71 (2.3\%) & 499 (16.5\%) & 3,023 \\
\midrule
Total & 10,647 (76.7\%) & 2,113 (15.2\%) & 457 (3.3\%) & 276 (2.0\%) & 383 (2.8\%) & 3,229 (23.3\%) & 13,876 \\
\bottomrule
\end{tabular}%
}
\end{table}

\begin{figure}[p]
\centering
\textbf{Trump}\par\smallskip
\includegraphics[width=0.96\linewidth,trim=0 0 1320bp 0,clip]{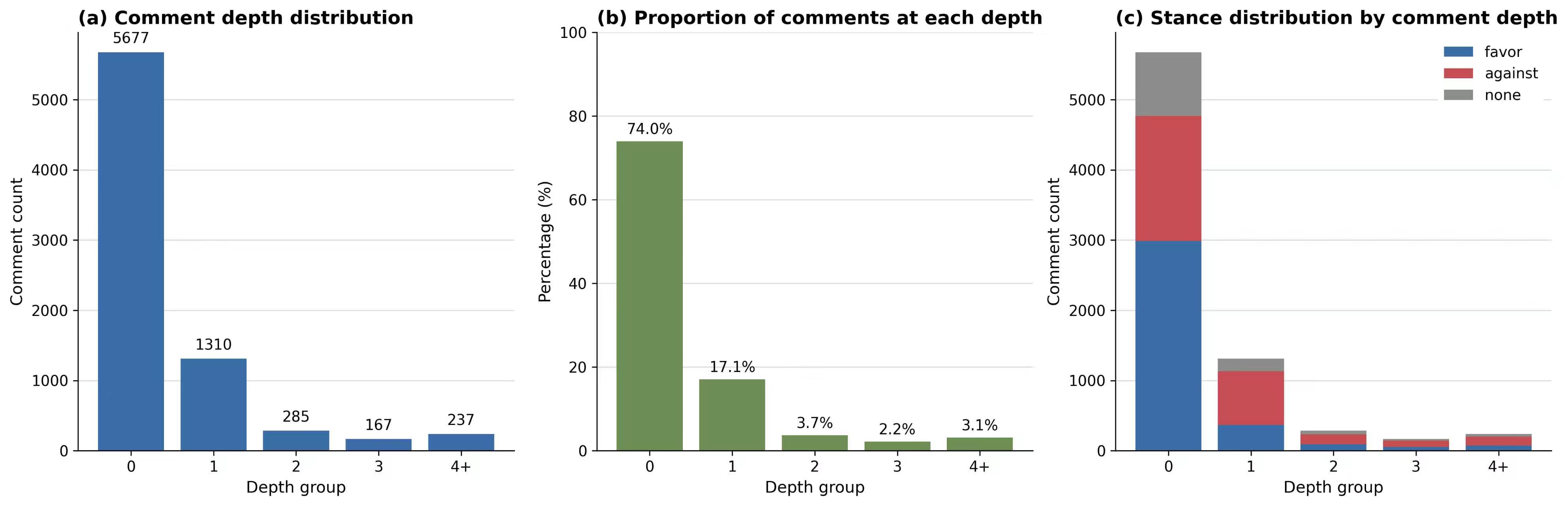}

\vspace{0.45em}
\textbf{Biden}\par\smallskip
\includegraphics[width=0.96\linewidth,trim=0 0 372bp 0,clip]{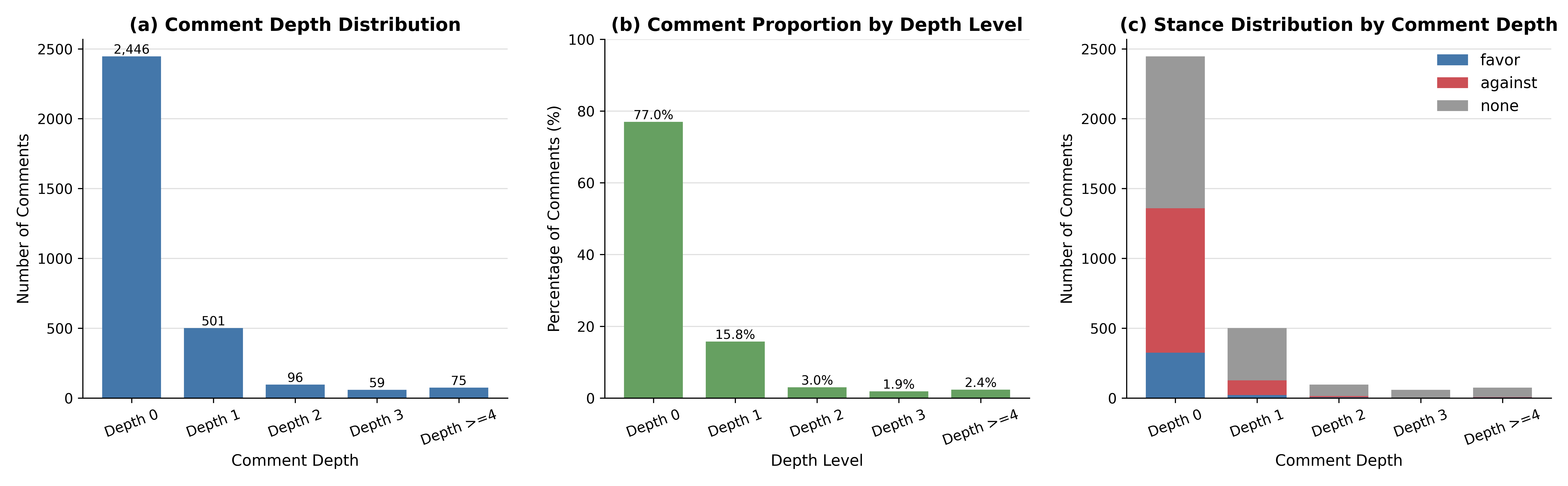}

\vspace{0.45em}
\textbf{Harris}\par\smallskip
\includegraphics[width=0.96\linewidth,trim=0 0 744bp 0,clip]{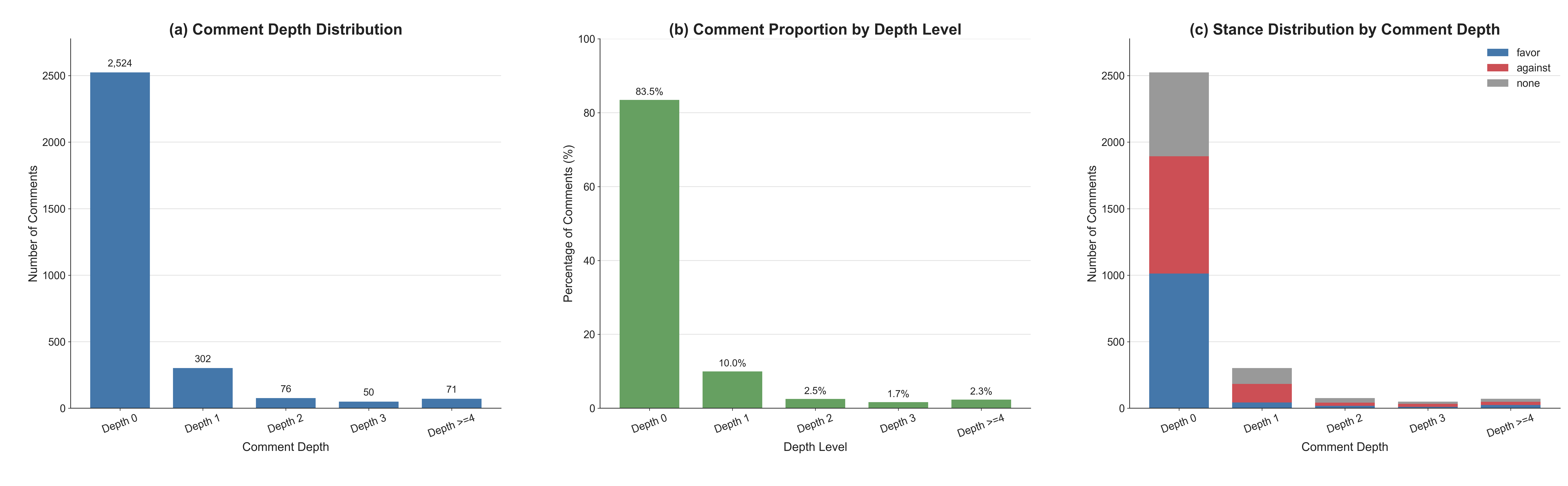}
\caption{\textbf{Reply-depth structure by target.} The original count and proportion panels are shown for Trump (\(n=7{,}676\)), Biden (\(n=3{,}177\)), and Harris (\(n=3{,}023\)). The stance-stack panels from the source graphics are omitted because exact stance-by-depth cells are unavailable for quantitative verification.}
\label{fig:analysis-depth}
\end{figure}

Figure~\ref{fig:analysis-depth} retains the count and proportion panels from the three target graphics; these panels agree with the marginal totals in Table~\ref{tab:analysis-depth}. The source graphics also contain stance stacks, but exact stance-by-depth cells are not available in the aggregate tables. Those stacks are therefore omitted rather than interpreted; a stance-by-depth analysis requires the contingency table.

\subsection{Bounded implications for model development and evaluation}
\label{subsec:analysis-modeling}

Three measured properties define concrete reuse constraints: one target contributes 63.4\% of videos, within-target class shares reach 55.2\% for Biden Against videos and 52.5\% for Biden None comments, and nested-comment prevalence ranges from 16.5\% to 26.0\%. These quantities reveal sampling and evaluation challenges within TikStance, but they are not model results and do not show that any architecture succeeds or fails. Table~\ref{tab:analysis-implications} expresses each pattern as an evaluation recommendation rather than as a performance claim.

\begin{table}[htbp]
\centering
\caption{\textbf{Evaluation implications supported by the descriptive aggregates.} Recommendations identify reuse designs motivated by measured composition and structure. No model, split, ablation, metric value, or statistical comparison has yet been evaluated.}
\label{tab:analysis-implications}
\small
\begin{tabularx}{\linewidth}{p{0.25\linewidth}X X}
\toprule
Observed pattern & Sample-bounded interpretation & Recommended reuse \\
\midrule
Trump supplies 102/161 videos (63.4\%) and 7,676/13,876 comments (55.3\%). &
Pooled training or scoring can be dominated by the largest retained target. &
Report per-target and macro-averaged metrics; consider leave-one-target-out evaluation; group all comments from a video in the same split. \\
\addlinespace
Video/comment class shares differ by target and level, including Biden's 55.2\% Against-video and 52.5\% None-comment shares. &
One global class prior would obscure target- and unit-specific imbalance. &
Report per-class metrics and confusion matrices separately for video-to-target and comment-to-target tasks; estimate weights from training data only. \\
\addlinespace
Nested replies comprise 3,229/13,876 comments (23.3\%), with target rates of 16.5--26.0\%. &
A nontrivial subset has parent or ancestor context, but deep bins are sparse. &
Compare comment-only, parent-aware, and ancestor-aware inputs when item-level data become available; report results by depth with exact cell sizes. \\
\bottomrule
\end{tabularx}
\end{table}

The analysis covers 161 videos, 13,876 comment-to-target labels, and five depth groups. It does not include a reconciled temporal series, stance-by-depth contingency table, inferential test, causal comparison, or model output. Within that boundary, TikStance motivates leakage-aware, target-stratified, multimodal, and hierarchy-sensitive evaluation; the performance of such designs remains to be established in separate experiments.

\FloatBarrier

\section{Discussion and Usage}
\label{sec:discussion}

\subsection{What the three-target design adds}

TikStance brings two lines of inquiry together: representing short-video discussion without discarding media and reply structure, and characterizing how the retained sample differs across political targets. Covering Biden and Harris alongside Trump changes the empirical picture rather than merely increasing the row count. Trump dominates the number of discussion units, but the smaller Biden and Harris subsets contain more retained comments per video. This mismatch means that a comment-weighted pooled analysis would give the targets different influence from a video-weighted analysis.

Post-filter stance composition also varies by target. Trump comments have a Favor plurality, Biden comments have a None majority, and the Harris Favor and Against totals are almost balanced. Video- and comment-level marginals differ in both direction and magnitude. Because they summarize different units, those contrasts are not transitions from creators to audiences and cannot show persuasion or echo-chamber formation. They do show that one pooled class prior is an inadequate description of the dataset.

Conversation structure supplies a second source of heterogeneity. Nearly one quarter of all retained comments are at depth 1 or greater, but the nested share is lowest for Harris. The sample therefore contains enough branch context to motivate ancestor-aware evaluation, while most comments remain top-level and the deepest reported bin combines depth 4 and all greater depths. The marginals do not reveal whether deeper replies are rebuttals, endorsements, or products of target-specific collection conditions.

\subsection{Evaluation implications and open questions}

Three practical evaluation choices follow from the observed distributions. First, target-dependent class composition makes an overall accuracy or micro-averaged score insufficient; per-target results and macro-averaged metrics are needed. Second, nested replies motivate comparisons between comment-only, ancestor-aware, video-aware, and combined inputs. Third, unequal target coverage motivates leave-one-target-out evaluation in addition to within-target splits. These are proposed protocols rather than completed model experiments.

The two stance levels offer a more distinctive question. A comment can oppose the political target promoted by a video, or endorse a target that the video criticizes. Quantitative claims about these configurations require item-level cross-tabulations of video stance and comment-to-target stance. Marginal totals cannot supply them. Once such joins are verified, the same framework can examine how cross-level relations vary by target and reply depth, with sparse cells reported rather than smoothed away.

Time remains useful but unresolved. The available temporal panels use incompatible denominators, timestamp definitions, and label-inclusion rules, including a Harris panel with 31 videos where the aggregate records support 30. They are therefore excluded from the comparative analysis. A regenerated series with one timestamp, timezone, cutoff, and inclusion rule could support temporal holdouts, but event markers alone would still not identify campaign effects, persuasion, or changes in public opinion.

\subsection{Usage Notes}
\label{sec:usage}

\subsubsection{Task formulations}

Once item-level joins and media availability are validated in a versioned release, the data model can support three core tasks. Video-level stance detection predicts Favor, Against, or None toward the designated target from the host-video content. Comment-to-target stance detection can be evaluated with comment text alone, with ancestor comments, with the host video, or with all contexts combined. A joint task can model the two levels together for records with complete item-level alignment and defined missing-value behavior. These tasks are proposed uses of the schema, not experiments reported in this paper.

Cross-target and temporal protocols are secondary task families. Leave-one-target-out evaluation asks whether a representation transfers across Trump, Biden, and Harris rather than memorizing target-specific expressions, an issue also foregrounded by target-diverse conversational work~\cite{ding2025zscsd}. Temporal evaluation should use one authoritative timestamp and a documented cutoff; the incompatible panels described above should not be merged into a benchmark split.

\subsubsection{Partitioning and reporting}

All records derived from one host video belong to the same data group. Train, development, and test partitions should therefore be created at the \texttt{video\_id} level, keeping the entire comment tree, its ancestors, and any media derivatives together. Random comment-level splitting would place shared context on both sides of an evaluation and would overstate generalization. With only 29 Biden and 30 Harris videos, repeated grouped folds may be more informative than a single 70/15/15 partition.

Report sample counts with every split and stratify results by target. Macro-F1 and per-class F1 are useful when label priors differ; micro-averaged scores should be accompanied by target-level results rather than treated as the sole summary. For context experiments, name the exact inputs available to each condition. “Multimodal” is too broad if one model sees only a transcript or sampled frames while another sees a decodable audio-video file.

\subsubsection{Label interpretation and missing context}

Each stance label is relational. \texttt{video\_stance} refers to the video's position toward its designated target and \texttt{stance\_target} to a comment's position toward that same target. Relevance is a separate decision. The label \texttt{None} denotes relevant content for which the codebook does not assign Favor or Against; it should not be collapsed automatically into generic sentiment neutrality. Consensus labels are human judgments, not error-free ground truth.

Ancestor context also needs explicit handling. A model should receive only ancestors that would be available at prediction time, in their documented order, with deleted or unavailable nodes represented consistently. If a released version lacks native media, a task must be described as ID-linked, transcript-based, or frame-based rather than full-video grounded. Dataset documentation should make these distinctions visible because composition, preprocessing, intended use, and maintenance are part of responsible reuse~\cite{gebru2021datasheets,pushkarna2022datacards}.

\subsubsection{Responsible reuse}

TikTok records are platform-dependent and may change after collection. Researchers should pin the dataset version, retain its manifest and checksums, and report whether missing media or stale links altered the analyzed sample. FAIR principles support persistent identification, rich metadata, explicit access conditions, and provenance, but FAIR does not mean unrestricted distribution~\cite{wilkinson2016fair}. Users should follow the applicable license or data-use agreement, honor withdrawal or takedown procedures, and avoid attempts to recover removed identifiers.

The sample must not be interpreted as representative of TikTok users, United States voters, or public opinion. Query terms, engagement thresholds, exposure, recommendation systems, deletion, and the collection window all shape which records are observed~\cite{olteanu2019social}. Stance labels should not be used to profile individuals or communities. Research questions about political communication should remain at the level authorized by the access conditions and supported by the sampling design.

\section{Conclusion}

TikStance defines target-conditioned discussion units for an analyzed sample of 161 political TikTok videos and 13,876 retained comments, with a schema for host-video context, parent-linked replies, and distinct video-to-target and comment-to-target judgments. Describing all three targets under common definitions reveals substantive heterogeneity: the dominant stance category changes by target and annotation level, while the prevalence of nested replies varies across the three subsets. These findings support target-stratified reporting, video-grouped partitions, and explicit context ablations as recommended evaluation choices. The current results remain descriptive of a query-based, post-filter sample, and public reuse depends on a versioned release with appropriate validation and access terms. Within those boundaries, TikStance provides a coherent foundation for studying how video, target, and conversational context interact in short-form political discussion.

\section*{Ethics Statement}
\label{sec:ethics}

Political social-media data concern people even when the source posts are publicly visible. Verbatim comments can be searchable; timestamps and reply structure can narrow a match; and video may reveal faces, voices, locations, or other contextual identifiers. Removing usernames and user IDs would reduce direct identifiability but would not make these records anonymous. Empirical work on acceptable uses of social-media data likewise cautions that sensitivity, social benefit, consent, and re-identification risk should be considered across the research lifecycle~\cite{hemphill2022sensitivity}.

An institutional ethics determination is not documented for this manuscript version; we therefore make no claim of exemption or of consent inferred from public visibility. Any data release is contingent on the responsible institution's determination and on documentation of collection scope, identifier handling, annotator exposure to sensitive content, and feasible withdrawal procedures. Data-card practice provides a useful structure for recording intended and out-of-scope uses~\cite{pushkarna2022datacards}.

The authority to collect a record and the authority to redistribute it are separate. Public visibility does not establish permission to redistribute native videos or verbatim comments, and source attribution alone does not resolve licensing~\cite{longpre2024licensing}. Access terms should distinguish metadata, comment text, media, and code; controlled or identifier-only access may be necessary for components that cannot lawfully be archived.

Privacy protection extends beyond direct identifiers. Exact timestamps and verbatim text can permit re-identification, while videos may expose faces, voices, or locations. Examples are therefore de-identified, and any access agreement should prohibit re-identification, individual profiling, harassment, and political targeting. Language-based privacy interventions may reduce disclosure risk but do not replace governance and human review~\cite{dou2024privacy}.

\FloatBarrier
\bibliographystyle{unsrt}
\bibliography{references}
\end{document}